\documentclass[journal]{IEEEtran}
\usepackage{cite}

\ifCLASSINFOpdf
  \usepackage[pdftex]{graphicx}
  
  \graphicspath{{../pdf/}{../jpeg/}{images/}}
  \DeclareGraphicsExtensions{.pdf,.jpeg,.png}
\else
  \usepackage[dvips]{graphicx}
  \DeclareGraphicsExtensions{.eps}
\fi
\usepackage{amsmath}
\usepackage{amssymb}
\usepackage{siunitx}
\usepackage{algorithm} 
\usepackage{algpseudocode} 

\usepackage{array}
\usepackage{multirow}

\usepackage{subcaption}

\usepackage{url}

\usepackage{xcolor, soul}
\sethlcolor{green}

\usepackage[pagebackref]{hyperref}
\usepackage{cleveref}
\usepackage{booktabs}
\usepackage{csquotes}

\title{MSSPlace: Multi-Sensor Place~Recognition with~Visual and~Text~Semantics}
\author{Alexander~Melekhin,
 Dmitry~Yudin,
 Ilia~Petryashin,
 Vitaly~Bezuglyj%

\thanks{This work was partially supported by the Analytical Center for the Government of the Russian Federation in accordance with the subsidy agreement (agreement identifier 000000D730321P5Q0002; grant No. 70-2021-00138).}
\thanks{Alexander~Melekhin, Ilia~Petryashin, Vitaly~Bezuglyj are with the Intelligent Transport Laboratory, Moscow Institute of Physics and Technology, Moscow Region, Dolgoprudny, 141700, Russia.}
\thanks{Dmitry Yudin is with Artificial Intelligence Research Institute (AIRI), Moscow, Russia and the Intelligent Transport Laboratory, Moscow Institute of Physics and Technology, Moscow Region, Dolgoprudny, 141700, Russia, e-mail: yudin@airi.net.}
}

\usepackage{tikz}
\usepackage{textcomp}
\newcommand\copyrighttext{%
  \footnotesize \textcopyright This work has been submitted to the IEEE for possible publication. Copyright may be transferred without notice, after which this version may no longer be accessible.
}
\newcommand\copyrightnotice{%
\begin{tikzpicture}[remember picture,overlay]
\node[anchor=south,yshift=10pt] at (current page.south) {\fbox{\parbox{\dimexpr\textwidth-\fboxsep-\fboxrule\relax}{\copyrighttext}}};
\end{tikzpicture}%
}

\begin{document}

\maketitle

\copyrightnotice

\begin{abstract}
    Place recognition is a challenging task in computer vision, crucial for enabling autonomous vehicles and robots to navigate previously visited environments. While significant progress has been made in learnable multimodal methods that combine onboard camera images and LiDAR point clouds, the full potential of these methods remains largely unexplored in localization applications. In this paper, we study the impact of leveraging a multi-camera setup and integrating diverse data sources for multimodal place recognition, incorporating explicit visual semantics and text descriptions. Our proposed method named MSSPlace utilizes images from multiple cameras, LiDAR point clouds, semantic segmentation masks, and text annotations to generate comprehensive place descriptors. We employ a late fusion approach to integrate these modalities, providing a unified representation. Through extensive experiments on the Oxford RobotCar and NCLT datasets, we systematically analyze the impact of each data source on the overall quality of place descriptors. Our experiments demonstrate that combining data from multiple sensors significantly improves place recognition model performance compared to single modality approaches and leads to state-of-the-art quality. We also show that separate usage of visual or textual semantics (which are more compact representations of sensory data) can achieve promising results in place recognition. The code for our method is publicly available: \url{https://github.com/alexmelekhin/MSSPlace}
\end{abstract}


%
\IEEEpeerreviewmaketitle

\section{Introduction}
\label{sec:intro}

\begin{figure}[t]
  \centering
  \includegraphics[width=\linewidth]{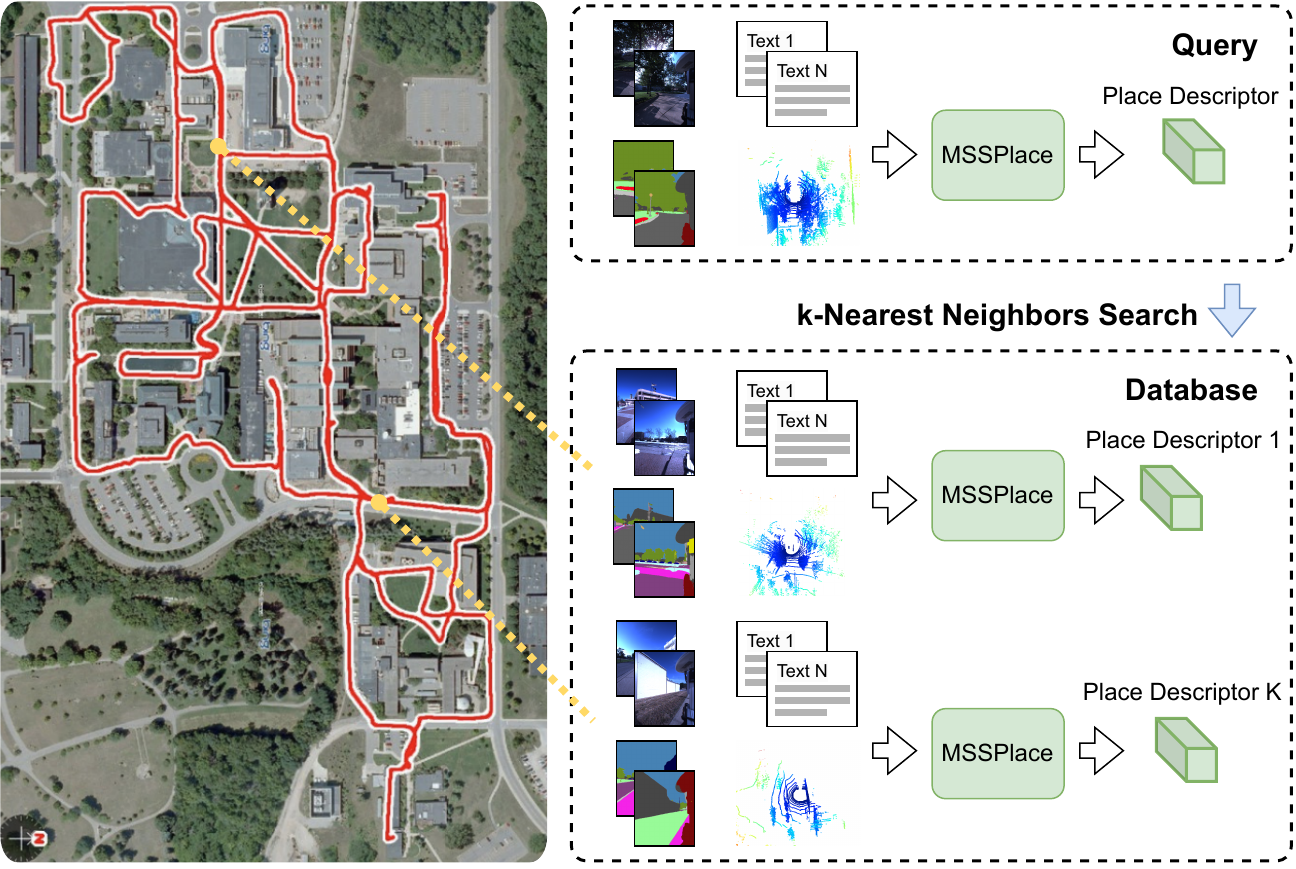}
  \caption{An overview of a multi-sensor place recognition pipeline with visual and text semantics. First, the input data is encoded into a query descriptor. A K-nearest neighbors search is then performed between the query and the database. Finally, the position of the closest database descriptor found is considered as the answer.}
  \label{fig:pr_overview}
\end{figure}

Place recognition is a fundamental component of the autonomous vehicle navigation methods, where known locations are identified by analyzing data from diverse sensors, including cameras and LiDARs. We define a \enquote{place} as a distinct physical location observed from a specific viewpoint \cite{garg_where_2021}. Recognizing these known places enables autonomous systems to achieve precise localization and efficient path planning, making robust place recognition crucial in various applications in robotics and intelligent transportation systems. The task is often formulated as a instance retrieval problem (see \Cref{fig:pr_overview} for details).

Single-modality methods often encounter limitations due to sensor-specific constraints. For example, camera-based methods \cite{arandjelovic_netvlad_2016,ali-bey_mixvpr_2023,berton_rethinking_2022,berton2023EigenPlacesTrainingViewpoint,keetha2024AnyLocUniversalVisual}, may lack spatial information, while LiDAR-based methods \cite{komorowski_minkloc3d_2021,komorowski_improving_2022,fan_svt-net_2022,ma_overlaptransformer_2022,luo_bevplace_2023,chen2023PTCNetPointWiseTransformer} may lack visual features. To overcome these limitations, multimodal methods \cite{xie_large-scale_2020, lu_pic-net_2020,komorowski_minkloc_2021,bernreiter_spherical_2021,pan_coral_2021,lai_adafusion_2022,yu_mmdf_2022} have been proposed, aiming to combine data from different sensors. By integrating information from multiple modalities, these methods address the drawbacks of single-modality approaches and enhance the overall performance. In addition, using data from multiple cameras simultaneously provides a more comprehensive representation of the environment, and can improve the performance even further.

In addition to incorporating data from multiple sensors, multimodal place recognition methods can benefit from explicit semantic masks \cite{acharya_single-image_2022,pramatarov_boxgraph_2022,kirilenko_vector_2022,ming_cgis-net_2022} and textual descriptions \cite{hong_textplace_2019,li_text-based_2020,kolmet_text2pos_2022,hettiarachchi_text_2022}. Semantic masks provide fine-grained information about different regions or objects present in the scene, enabling a more detailed understanding of the environment. Text descriptions associated with observed places may offer complementary contextual information, enhancing the discriminative power of the recognition system.  

Another significant benefit of incorporating text descriptions in multimodal place recognition is the ability to describe recognized places in a human-readable form. Text descriptions provide a natural language representation of the environment, facilitating seamless communication and interaction between humans and robots \cite{kolmet_text2pos_2022, text2pos_transf}. By associating text descriptions with recognized places, autonomous agents can provide informative and contextually relevant descriptions to humans, enhancing the interpretability and explainability of the system. This capability becomes particularly valuable in human-robot interactions, as it fosters collaboration, user trust, and intuitive navigation experiences.

In our analysis, we look at how each part -- multiple sensors, explicit semantic masks, and text descriptions -- contributes separately. This helps us understand their individual roles in place recognition methods. By examining each modality closely, we refine our understanding, paving the way for more informed multimodal approaches that improve performance and robustness in place recognition.

The main goal of this research is to quantitatively evaluate the performance improvements achieved by incorporating and assessing the impact of individual modalities, including LiDAR point clouds, data from multiple cameras, semantic masks, and text descriptions, in the context of multimodal place recognition. Our objective is to closely examine the performance of each modality independently and in diverse combinations using the late fusion approach, with a focus on enhancing accuracy, robustness, and overall system performance in place recognition. Through extensive experiments on widely-used open datasets, such as the Oxford RobotCar \cite{maddern_1_2017,maddern_real-time_2020} and NCLT \cite{carlevaris-bianco_university_2016}, we aim to provide valuable insights into the benefits of incorporating these modalities. This exploration paves the way for the development of more effective multimodal place recognition techniques. 

Our contribution may be summarized as follow:

\begin{itemize}
    \item We extend the Oxford RobotCar and NCLT datasets by incorporating semantic segmentation masks and text descriptions generated using OneFormer \cite{oneformer_23} and MiniGPT-4 \cite{zhu_minigpt-4_2023} models, facilitating further research in multimodal place recognition.
    \item We conduct a thorough investigation into the performance of each modality separately, utilizing well-established neural network architectures with proven effectiveness in place recognition tasks. This allows us to understand the individual strengths and limitations of camera, LiDAR, semantic masks, and text descriptions.
    \item We develop a modular neural network method, named MSSPlace, that combines data from different modalities using the late fusion technique. It can be easily customized by incorporating any suitable model for each modality, providing flexibility and adaptability in multimodal place recognition scenarios.
\end{itemize}

\setcounter{section}{1}
\section{Related Work}
\label{sec:related}

In this section, we provide a brief overview of the related work in multimodal place recognition and methods that leverage image and text semantics.

\subsection{Single-modality place recognition}

Image-based place recognition has been extensively studied, with notable contributions such as NetVLAD \cite{arandjelovic_netvlad_2016}, which introduced a learnable VLAD layer for feature aggregation. Recent methods like CosPlace \cite{berton_rethinking_2022}, EigenPlaces \cite{berton2023EigenPlacesTrainingViewpoint}, and MixVPR \cite{ali-bey_mixvpr_2023} have advanced the field by leveraging large datasets and innovative architectures, including transformers.
In the pursuit of generalization, AnyLoc \cite{keetha2024AnyLocUniversalVisual} employ pre-trained models like DINOv2 \cite{oquab2023dinov2} to extract features and utilize a VLAD layer for aggregation.

LiDAR data contains rich geometric information crucial for place recognition. Approaches like MinkLoc3D \cite{komorowski_minkloc3d_2021} and its modification \cite{komorowski_improving_2022} voxelize input point clouds and process them with ResNet-like architectures to generate a single descriptor. SVT-Net \cite{fan2022SVTNetSuperLightWeight} employs a similar architecture augmented with attention modules.
Alternatively, methods like OverlapTransformer \cite{ma_overlaptransformer_2022} and BEVPlace \cite{luo_bevplace_2023} project point clouds onto range images and bird's-eye-view images, respectively. PTC-Net \cite{chen2023PTCNetPointWiseTransformer} integrates point-wise and voxel-based approaches using transformer architecture, combining their respective advantages.

\subsection{Multimodal place recognition}  

Although the idea of combining the two modalities may seem obvious, the first work in this area has only appeared relatively recently. Xie et al. \cite{xie_large-scale_2020} introduced a trimmed VLAD module for global point cloud descriptor extraction, along with ResNet50 for image descriptor extraction.
Another method, PIC-net \cite{lu_pic-net_2020}, explored the use of attention modules. It leveraged ResNet50 for image feature extraction and PointNet \cite{qi_pointnet_2017} or LPD-Net \cite{liu_lpd-net_2019} for point cloud feature extraction. The features are then passed through a novel Spatial Attention NetVLAD layer, a channel attention layer, and a global attention layer.
In \cite{zywanowski_comparison_2020}, a camera image is concatenated with a LiDAR intensity image, and a VGG16-based feature extractor \cite{olid_single-view_2018} is used for further processing.
The MinkLoc++ method \cite{komorowski_minkloc_2021} employs MinkLoc3D \cite{komorowski_minkloc3d_2021} and ResNet18 to extract feature maps for point clouds and images, respectively. These feature maps are then concatenated after applying a GeM pooling layer \cite{radenovic_fine-tuning_2019}.
Bernreiter et al. \cite{bernreiter_spherical_2021} project multimodal input onto a sphere and process it with a spherical CNN to learn a unique embedding for end-to-end Place Recognition.
CORAL \cite{pan_coral_2021} utilizes a dense elevation image representation of LiDAR data and project map camera image features onto it. The fused multimodal data is then passed through a NetVLAD \cite{arandjelovic_netvlad_2016} layer for local feature aggregation and to obtain a global descriptor.
AdaFusion \cite{lai_adafusion_2021} utilizes CNN architecture with multi-scale attention to learn weights for both image and point cloud features to better utilize their relationship.
MMDF \cite{yu_mmdf_2022} proposes a multi-modal fusion module for cross-scene place recognition. That module enhances the point cloud features from PointNet with image features from simple CNN network.


\subsection{Semantic masks for place recognition}  

Although semantic masks may contain less information compared to images or LiDAR scans, they have the advantage of being invariant to visual scene changes such as illumination and seasonal variations. This desirable property makes them well suited for generating stable scene descriptors. Semantic masks can be used in two ways: first, implicitly, to form semantically meaningful visual descriptors during model training \cite{larrson_finegrained_2019, dasgil, multiatt_2022, mingCGiSNetAggregatingColour2022} and second, directly, to construct descriptors. Furthermore, the semantics can be used to create individual embeddings \cite{hybrid_imret_22, robust_hybrid_22, global_inference_22, object_matter_22, kirilenko_vector_2022} or to enrich image descriptors from the camera \cite{LOST_2018, context_agg_2020}.

Our approach follows closely that of \cite{robust_hybrid_22}, where we use semantic masks as an independent modality separate from the images.

\subsection{Text semantics for place recognition}  

The utilization of text modality in place recognition methodologies has emerged relatively recently \cite{gpt_clip_generation}. The primary focus of studies in this area revolves around establishing alignment between textual descriptors and conventional LiDAR point cloud or image descriptors \cite{text2pos_transf, kolmet_text2pos_2022}. These methodologies aim to explore the feasibility of identifying similar locations based on human language queries. However, it is important to note that the textual descriptions employed in these approaches are formalized and derived from semantic scene descriptions, thereby creating a distinction from natural language. Furthermore, these methodologies have yet to consider the potential of integrating text descriptors with other feasible descriptors.

Moreover, this field of study investigates techniques for accurately localizing objects described in natural language within a given scene \cite{lin2023wildrefer}. Additionally, it explores methodologies for detecting and utilizing text present in images as an additional scene descriptor \cite{hong_textplace_2019, mti6110102, ocr_retrieve}.

\section{Methodology}
\label{sec:methodology}

\begin{figure*}[t]
  \centering
  \includegraphics[width=0.7\linewidth]{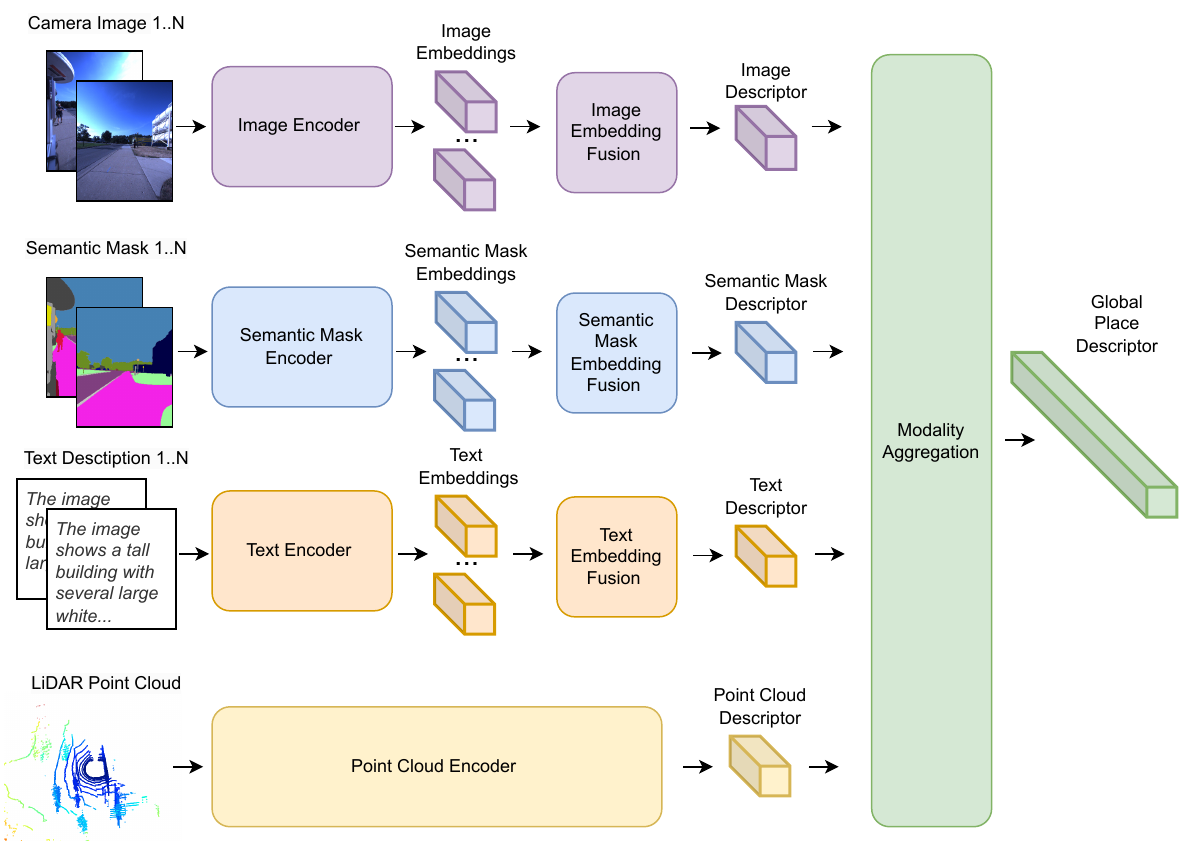}
  \caption{High-level overview of the proposed multimodal MSSPlace method. The method has a modular architecture and consists of four branches: the Image Encoder, Semantic Masks Encoder, Text Encoder, and Point Cloud Encoder. Each branch encodes the input data into a descriptor, capturing the essential information specific to its respective modality. Subsequently, a descriptor aggregation step is performed to combine these individual descriptors and obtain the global place descriptor, which represents the comprehensive characteristics of the vehicle place.}
  \label{fig:method}
\end{figure*}

\begin{figure*}[t]
  \centering
  \begin{subfigure}{0.6\linewidth}
    \centering
    \includegraphics[height=3cm]{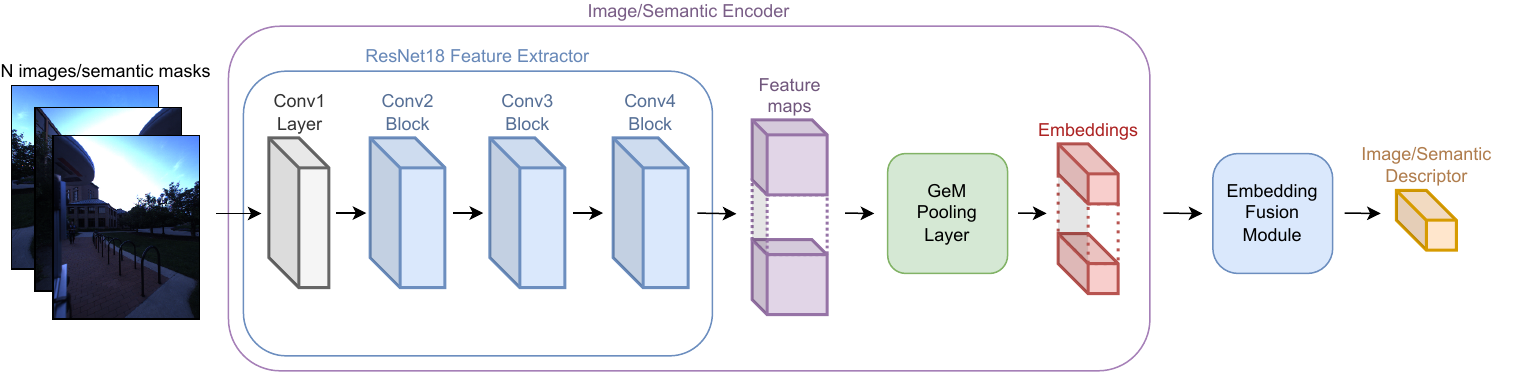}
    \caption{}
    \label{fig:image-semantic-encoder-a}
  \end{subfigure}
  \hfill
  \begin{subfigure}{0.3\linewidth}
    \centering
    \includegraphics[height=3cm]{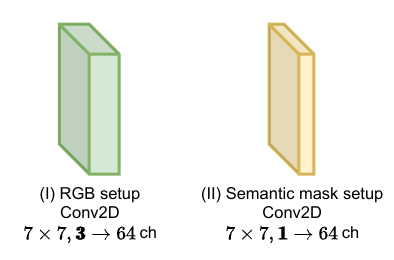}
    \caption{}
    \label{fig:image-semantic-encoder-b}
  \end{subfigure}
  \caption{Image and semantic encoder scheme. Both encoders have the same architecture based on ResNet18 with difference in input channels number: (a) ResNet18-based encoder architecture; (b) Different setups of 1st ResNet convolutional layer: (I) - for RGB images, (II) - for semantic masks.}
  \label{fig:image-semantic-encoder}
\end{figure*}

\begin{figure*}[t]
  \centering
  \begin{subfigure}{0.55\linewidth}
    \centering
    \includegraphics[height=1.8cm]{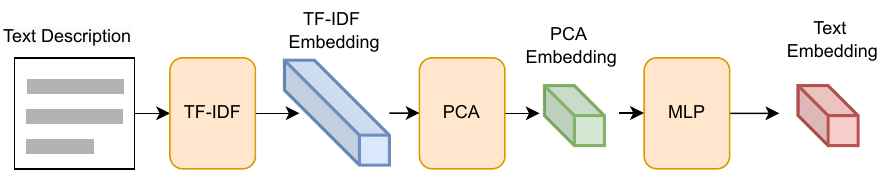}
    \caption{}
    \label{fig:text_encoders-a}
  \end{subfigure}
  \hfill
  \begin{subfigure}{0.44\linewidth}
    \centering
    \includegraphics[height=1.8cm]{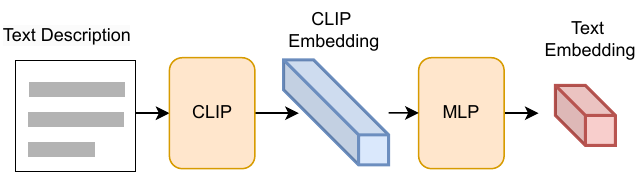}
    \caption{}
    \label{fig:text_encoders-b}
  \end{subfigure}
  \caption{Text encoders: (a) TF-IDF-based; (b) CLIP text encoder-based. }
  \label{fig:text_encoders}
\end{figure*}

The place recognition task is often formulated as an instance retrieval task: we want to find the most similar element in the \textit{database} for the given \textit{query}. Lets assume that we have $P_q$ as our query place and $\mathcal{P}_{db} = \{P_{db_1}, ..., P_{db_n}\}$ as a set of all places in database.  We want to learn mapping function $F(\cdot)$ so that it will produce descriptor vector $p$ of desirable dimension $D$ for given input data: $p = F(P)$, where $p \in \mathbb{R}^D$. We use the Euclidean distance as a measure of similarity for two descriptors. So the distance between positive pair $(p_q, p_{db_{pos}})$ (representing the same place) should be small, while the distance between negative pair $(p_q, p_{db_{neg}})$ (representing the different places) should be large:

\begin{equation}
    || p_q - p_{db_{pos}} ||_2 \to 0, \quad || p_q - p_{db_{neg}} ||_2 \to \infty.
\end{equation}

In the proposed method, the descriptor vector $p$ is defined as follows:
    $p = A(p_{mc}, p_{ms}, p_{mt}, p_{l}),$
where $A$ represents the descriptor aggregation function. The descriptor $p_{mc}$ is obtained by fusing the individual camera descriptors ${p_{c1}, ..., p_{cn}}$ using a fusion operation. Similarly, the descriptors $p_{ms}$ and $p_{mt}$ are obtained by fusing the individual descriptors from multiple semantic masks and multiple text descriptions, respectively.


Metric learning techniques are usually used to train the neural networks for the place recognition task. We use a commonly adopted triplet margin loss \cite{hermans_defense_2017}. It packs input elements into triplets $(p_{anchor}, p_{positive}, p_{negative})$. The loss is defined as follows:

\begin{equation}
    \mathcal{L}_{triplet} = [d_{ap} - d_{an} + m]_+,
\end{equation}
where $[x]_+$ is the $max(0, x)$ operation; $d_{ap}$ and $d_{an}$ are the distances between anchor-positive and anchor-negative pairs of descriptors in the input triplet; and $m$ is the margin hyperparameter.

\subsection{MSSPlace method}

As illustrated in \Cref{fig:method}, the proposed method performs a late fusion of descriptors from multiple modalities, which are generated by independent branches. The Modality Aggregation module performs concatenation of all input descriptors. 

\paragraph{Image Encoder module}
We employ the ResNet18 architecture with GeM pooling to extract individual image embeddings. These embeddings are subsequently passed through the Embedding Fusion module, which combines them to generate the final image descriptor. We explored different options for combining the individual image embeddings using various Embedding Fusion modules. These options included Concatenation, Addition, MLP, and Self-Attention. Refer to \Cref{fig:image-semantic-encoder} for an illustration of the architecture.


\paragraph{Semantic Mask Encoder module}
As mentioned above, our semantic and image encoders share the same architecture based on ResNet18 (\Cref{fig:image-semantic-encoder-a}), similar to the approach used in \cite{robust_hybrid_22}. By using a convolutional network to derive semantic features, we can extract more complex features than with the histogram method \cite{hybrid_imret_22}, while also allowing for easy modification and integration with the late fusion approach widely used in \cite{komorowski_minkloc_2021}. In addition, using the same network as for images makes it easier to experiment with feature fusion. To handle semantic masks (monochannel image), we've adapted ResNet layer 1 (\Cref{fig:image-semantic-encoder-b}).

VGG16 was used as the underlying architecture in \cite{robust_hybrid_22}, but based on the excellent results demonstrated in \cite{komorowski_minkloc_2021} using ResNet18, it seems redundant to use a more complex architecture.

\paragraph{Text Encoder module}
We use TF-IDF as a baseline for feature extraction from text, without imposing any restrictions on the vocabulary size. To reduce the dimensionality of the TF-IDF features, we employ PCA to extract 128 components. We then construct a two-layer MLP on top of the PCA embeddings, with the number of neurons in the output layer set to 128.

For comparison, we also test the CLIP \cite{openai_clip} text encoder as a basic model for feature extraction from text. We use both the base and large implementations of CLIP, which have 512 and 768 embedding sizes, respectively. On top of the CLIP embeddings, we construct a two-layer MLP with the number of output layer neurons set to 256.

The two approaches are illustrated in \Cref{fig:text_encoders}. To combine embeddings from multiple texts, we use either addition or concatenation. In the multimodal setup we use only addition. 


\paragraph{Point Cloud Encoder module}
We employ the MinkLoc3D architecture \cite{komorowski_minkloc3d_2021} as the foundation for our Point Cloud Encoder, incorporating modifications introduced in MinkLoc++ \cite{komorowski_minkloc_2021}. This architecture builds upon a simple ResNetFPN framework, enhanced by the integration of ECA layers \cite{wang_eca-net_2020} and GeM pooling \cite{radenovic_fine-tuning_2019}. The MinkLoc3D architecture serves as a powerful feature extraction backbone for our point cloud data. By leveraging the ResNetFPN structure, it captures multi-scale representations of the input point cloud, enabling the extraction of rich and discriminative features. The incorporation of ECA layers further enhances the model's ability to capture long-range dependencies and capture more contextual information within the point cloud. Additionally, GeM pooling provides an effective means of aggregating the learned features, facilitating the generation of compact and informative descriptors.

\begin{figure}[t]
  \centering
  \includegraphics[width=0.99\linewidth]{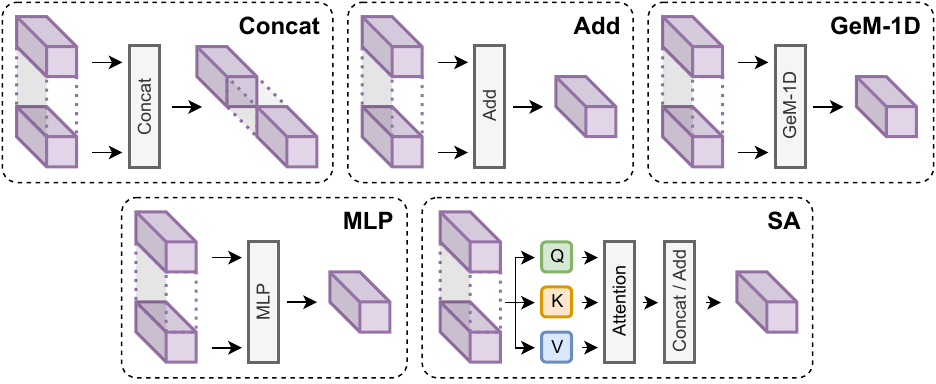}
  \caption{Embedding Fusion modules}
  \label{fig:image_fusion_modules}
\end{figure}

\paragraph{Embedding Fusion module}
We explore different options for combining the individual embeddings using various Embedding Fusion modules. Figure \ref{fig:image_fusion_modules} illustrates the details of module implementations. The Concatenation Fusion module performs concatenation of all input embeddings. The Addition Fusion module performs the addition of all input embeddings. The MLP Fusion concatenates all input embeddings, passes them through 2-layer MLP with $dim_{input} = dim_{hidden}$, GELU activation between layers and output dimension equal to either $256$ or $512$. The Self-Attention Fusion module implements self-attention between input embeddings and concatenates or adds them to produce the final descriptor. The GeM-1D module stacks input embeddings and performs one-dimensional GeM pooling to produce the final descriptor.

\subsection{Dataset preparation}


We utilize the preprocessed version of the Oxford RobotCar dataset \cite{maddern_1_2017, maddern_real-time_2020}, as introduced in the PointNetVLAD method \cite{uy_pointnetvlad_2018}. Following the approach of MinkLoc++ \cite{komorowski_minkloc_2021}, we extend the dataset with camera images, extracting only one image per LiDAR point cloud. However, we observed missing frames for the \textit{mono\_left}, \textit{mono\_rear}, and \textit{mono\_right} cameras, which we exclude from the dataset. This results in a smaller dataset size, which should be considered when comparing with other methods. For the NCLT dataset \cite{carlevaris-bianco_university_2016}, we adopt the 10 tracks and split described in the AdaFusion paper \cite{lai_adafusion_2022}. However, since we do not have access to the original author's code for preprocessing, it is important to note that the resulting split may differ from the original.

To use text and semantic modalities, we extend Oxford RobotCar and NCLT datasets with text descriptions and semantic segmentation masks.

To perform semantic segmentation, we use state-of-the-art OneFormer model \cite{oneformer_23} pretrained on the Mapillary dataset \cite{mapillary_17}. The model is capable of working in semantic, instance and panoptic segmentation modes, but for our datasets we focus only on semantic segmentation, with a total of 65 semantic classes. We obtain semantic masks for each camera in both Oxford RobotCar \cite{uy_pointnetvlad_2018} (4 cameras) and NCLT dataset \cite{carlevaris-bianco_university_2016} (5 cameras).

\begin{figure*}[t]
  \centering
  \includegraphics[width=0.90\linewidth]{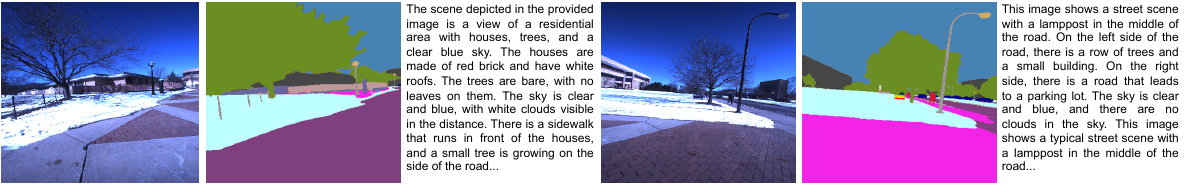}
  \caption{Extension of NCLT dataset with semantic masks and text descriptions}
  \label{fig:extended_nclt}
\end{figure*}

To obtain text descriptions, we use MiniGPT-4 \cite{zhu_minigpt-4_2023} with a checkpoint based on Vicuna-13B. We use the "Describe this scene" prompt to generate image descriptions, without any pre-processing of the input images or post-processing of the generated texts. Examples of generated descriptions are shown in \Cref{fig:extended_nclt}.

\section{Experiments}
\label{sec:experiments}

\subsection{Training details}

The training procedure is mostly inherited from MinkLoc++ \cite{komorowski_minkloc_2021} method.The models are trained for 80 epochs with Adam optimizer with learning rate scaled by 0.1 after 40 and 60 epochs. Initial learning rate was set to 0.001 for point cloud and semantic mask encoders and 0.0001 for image and text encoders. All experiments were performed on an NVIDIA Tesla V100 32GB.

\paragraph{Batch hard negative mining}
To mine the triplet, we find the hardest positive and hardest negative samples for each element in the batch.

\paragraph{Dynamic batch sizing}
The dynamic batch sizing procedure solves the problem of early training collapse. The training starts with the initial batch size equal to 16. Then, if the ratio of zero-loss elements in the epoch is below 0.7, the batch size increases by 1.4. The maximum batch size is 128. Combined with the batch hard negative mining, it allows a gradual increase in the number of difficult samples in the batch.

\paragraph{Descriptor size}
We set the size of the output embedding for each Encoder to 256. In multi-sensor experiments, we use the Addition Fusion module for the Embedding Fusion block, and Concatenation for the Modality Aggregation block. Therefore, the size of the final Global Place Descriptor is equal to $N \times 256$, where $N$ is the number of modalities used in the model.

\subsection{Evaluation metrics}
\label{subsec:eval_metrics}

We employ a commonly adopted evaluation procedure. The query is considered localized if at least one of the top $N$ retrieved database frames is within $d = 25$ meters of the ground truth position of the query. We report the \textit{Average Recall @ 1} and \textit{Average Recall @ 1\%} metrics.

\subsection{Experiments with multiple images}
\label{subsec:image_experiments}

In our experiments with multiple camera images, we investigate various combinations of cameras (see \Cref{tab:camera_setups} for details) and different fusion modules. \Cref{tab:camera_ablation} provides the results of the experiments and their corresponding descriptions: \textit{Concat} refers to the concatenation of embeddings, \textit{Add} represents the addition of embeddings, \textit{GeM-1D} refers to the GeM pooling module, \textit{MLP$_{dim}$} signifies the concatenation of embeddings followed by a 2-layer MLP with output dimension equal to \textit{dim}. Additionally, \textit{SA} denotes Self-Attention between the input embeddings. For the \textit{all cameras} experiments we use only concatenation, addition and GeM pooling, as it showed the best quality for 2 cameras.

For the two-camera configuration, we explore two scenarios: one with front and back cameras, and another with left and right cameras. The hypothesis was that cameras positioned to the sides of the car might capture more useful information, such as buildings along the road. However, contrary to the hypothesis, the configuration with front and back cameras demonstrated the best result.

\begin{table}[t]
  \caption{Abbreviations for combinations of cameras serving as sources of images, semantic masks, and text descriptions.}
  \label{tab:camera_setups}
  \centering
  \scriptsize
  \begin{tabular}{lcc}
    \toprule
    Abbreviation & RobotCar & NCLT \\
    \midrule
    F & stereo\_centre & Cam5 \\
    F+B & stereo\_centre, mono\_rear & Cam5, Cam2 \\
    L+R & mono\_left, mono\_right & Cam1, Cam4 \\
    \multirow{2}{*}{A} & stereo\_centre, mono\_left, & Cam1, Cam2, Cam3, \\
     & mono\_rear, mono\_right & Cam4, Cam5 \\
    \bottomrule
  \end{tabular}
\end{table}

Experiments with MLP prove that adding extra learnable parameters leads to overfitting and worse generalization. The same, but to a lesser extent, can be said about Self-Attention. Adding SA layer resulted in nearly the same performance.

The most interesting conclusions can be drawn from the fact that for Oxford RobotCar the best quality was achieved by the concatenation method, and for NCLT by the addition method. This discrepancy can be attributed to the rotation invariance of the addition method. In the case of RobotCar, where the car consistently follows the same trajectory in a fixed direction, instances where the query and database share a location but exhibit an expanded view are rare. Conversely, in NCLT, where the robot follows a slightly less rigid trajectory, situations often arise where the robot turns differently while being in the same place. This characteristic makes the rotation-invariant addition method more effective in handling such scenarios, contributing to its superior performance on the NCLT dataset.

\begin{table}[t]
  \caption{Results of the experiments with multiple camera images. We use following abbreviations: F -- image for front camera, F+B -- images for front and back cameras, L+R -- images for left and right cameras, A -- images for all cameras. The best result in each section is in bold, the second is underlined.}
  \label{tab:camera_ablation}
  \centering
  \scriptsize
  \begin{tabular}{lccccc}
    \toprule
    \multirow{2}{*}{Cams} & \multirow{2}{*}{Fusion method} & \multicolumn{2}{c}{RobotCar}  &    \multicolumn{2}{c}{NCLT}  \\
     &   &   AR@1  &   AR@1\%               &   AR@1   &   AR@1\%   \\
    \midrule
    F & -              &         84.56  &  93.62  &  65.88  &  76.88 \\
    \midrule
    F+B & Add & \underline{88.37} & \underline{96.25} & \textbf{75.20} & \textbf{85.52} \\
    F+B & Concat & \textbf{92.82} & \textbf{97.78} & 72.02 & 79.79 \\
    F+B & GeM-1D & 87.96 & 95.60 & \underline{74.70} & \underline{85.36} \\
    F+B & MLP$_{512}$ & 80.50 & 93.00 & 62.55 & 78.18 \\
    F+B & MLP$_{256}$ & 78.04 & 92.31 & 59.90 & 76.42 \\
    F+B & SA+Add & 83.28 & 94.59 & 72.17 & 84.21 \\
    F+B & SA+Concat & 81.56 & 93.75 & 71.62 & 83.74 \\
    \midrule
    L+R & Add & \underline{82.85} & \underline{93.65} & \underline{75.41} & \underline{86.45} \\
    L+R & Concat & \textbf{86.93} & \textbf{95.07} & 70.57 & 79.26 \\
    L+R & GeM-1D & 82.64 & 93.71 & \textbf{75.45} & \textbf{86.64} \\
    L+R & MLP$_{512}$ & 73.98 & 89.63 & 57.89 & 74.67 \\
    L+R & MLP$_{256}$ & 73.83 & 90.45 & 57.78 & 74.25 \\
    L+R & SA+Add & 78.93 & 92.22 & 72.17 & 84.78 \\
    L+R & SA+Concat & 76.81 & 91.54 & 70.36 & 83.11 \\
    \midrule
    A & Add & 89.84 & \underline{97.23} & \textbf{88.50} & \textbf{94.75} \\
    A & Concat & \textbf{95.52} & \textbf{98.70} & 77.03 & 83.16 \\
    A & GeM-1D & \underline{90.33} & 97.15 & \underline{88.15} & \underline{94.50} \\
    \bottomrule
  \end{tabular}
\end{table}

\subsection{Experiments with semantic masks}
\label{subsec:semantic_experiments}

In our semantic experiments, we assess the effectiveness of place recognition solely using the semantic modality. The results of these experiments are presented in \Cref{tab:semantic_ablation}.

We employ the same configurations as described in \cref{subsec:image_experiments}. The results on the NCLT dataset differ from the experiments with cameras. The GeM pooling showed the best performance in all configurations, indicating that learnable pooling may provide benefits over simple addition. In general, the results show that even in the absence of rich visual information, place recognition remains effective when using only semantic masks as input.

\begin{table}[t]
  \caption{Results of the experiments with semantic masks. We use the following abbreviations: F -- semantic mask for front camera, F+B -- semantic masks for front and back cameras, L+R -- semantic masks for left and right cameras, A -- semantic masks for all cameras. 
  }
  \label{tab:semantic_ablation}
  \centering
  \scriptsize
  \begin{tabular}{lccccc}
    \toprule
    \multirow{2}{*}{Cams} & \multirow{2}{*}{Fusion method} & \multicolumn{2}{c}{RobotCar}  &    \multicolumn{2}{c}{NCLT}  \\
     &   &   AR@1  &   AR@1\%               &   AR@1   &   AR@1\%   \\
    \midrule
    F & - & 84.95 & 93.67 & 60.87 & 71.86 \\
    \midrule
    F+B & Add & 86.17 & 94.39 & \underline{71.30} & \underline{82.29} \\
    F+B & Concat & \textbf{90.05} & \textbf{96.25} & 68.36 & 76.06 \\
    F+B & GeM-1D & \underline{89.23} & \underline{95.73} & \textbf{74.65} & \textbf{83.96} \\
    F+B & MLP$_{512}$ & 72.80 & 87.58 & 54.92 & 71.92 \\
    F+B & MLP$_{256}$ & 80.64 & 91.62 & 50.15 & 69.24 \\
    F+B & SA+Add & 80.08 & 91.96 & 66.21 & 79.40 \\
    F+B & SA+Concat & 80.37 & 92.04 & 64.90 & 78.32 \\
    \midrule
    L+R & Add & 67.51 & 82.53 & \underline{68.70} & \underline{81.25} \\
    L+R & Concat & \textbf{72.90} & \textbf{86.00} & 63.78 & 73.66 \\
    L+R & GeM-1D & \underline{68.65} & \underline{83.52} & \textbf{72.00} & \textbf{84.06} \\
    L+R & MLP$_{512}$ & 60.18 & 77.88 & 54.08 & 71.00 \\
    L+R & MLP$_{256}$ & 59.38 & 77.57 & 56.52 & 73.31 \\
    L+R & SA+Add & 60.83 & 78.76 & 61.64 & 77.49 \\
    L+R & SA+Concat & 61.01 & 79.29 & 65.28 & 79.76 \\
    \midrule
    A & Add & 81.51 & 92.38 & \underline{83.90} & \underline{91.17} \\
    A & Concat & \textbf{91.35} & \textbf{96.94} & 72.49 & 79.00 \\
    A & GeM-1D & \underline{87.44} & \underline{95.01} & \textbf{87.84} & \textbf{93.43} \\
    \bottomrule
  \end{tabular}
\end{table}

\subsection{Experiments with text descriptions}
\label{subsec:text_experiments}

In our experiments focused on place recognition using only the textual modality, we employ the same sensor configurations as described in \cref{subsec:image_experiments}. Two methods of combining multiple embeddings from different cameras are explored: concatenation and addition. The results are presented in \Cref{tab:text_ablation}.


Our experiments highlight that the superior quality is achieved when utilizing CLIP-large as the feature extractor, attributing this to the higher quality features it produces.

Experiments on both datasets indicate that incorporating information from the back camera yields only marginal quality improvement, as it does not contribute substantial semantic information. In contrast to the results presented in \cref{subsec:image_experiments} and \cref{subsec:semantic_experiments}, the simultaneous use of text descriptions from side cameras outperformed the combination of front and back camera information. This observation stems from the side cameras capturing more specific scene details, subsequently reflected in the descriptions generated by MiniGPT-4. The most favorable results were obtained when utilizing information from all cameras, enabling comprehensive scene descriptions.

Notably, the quality of results on the RobotCar dataset was significantly inferior to that on the NCLT dataset. Further analysis of the text descriptions revealed that RobotCar descriptions utilized approximately 30\% fewer unique words compared to NCLT. This disparity indicates a smaller variety of objects describing the scene in RobotCar. The discrepancy is attributed to the homogeneous urban environment in RobotCar images, whereas NCLT encompasses diverse locations, including indoor settings. Additionally, the larger size of RobotCar poses additional challenges to the recognition task.

\begin{table}[t]
  \caption{Results of the experiments with text descriptions. We use the following abbreviations: F -- text description for front camera, F+B -- text descriptions for front and back cameras, L+R -- text descriptions for left and right cameras, A -- text descriptions for all cameras. 
  }
  \label{tab:text_ablation}
  \centering
  \scriptsize
  \begin{tabular}{lccccc}
    \toprule
    \multirow{2}{*}{Cams} & \multirow{2}{*}{Method} &        \multicolumn{2}{c}{RobotCar}  &            \multicolumn{2}{c}{NCLT}  \\
        &   &            AR@1  &            AR@1\% &             AR@1  &           AR@1\% \\
    \midrule
    F       & TF-IDF           &            2.64 &             7.27 &            11.44 &            24.49 \\
    F       & CLIP-B               &            \underline{2.71} &             \underline{7.45} &            \underline{13.99} &            \underline{27.67} \\
    F       & CLIP-L                 &            \textbf{2.94} &             \textbf{7.99} &            \textbf{14.60} &            \textbf{29.03} \\
    \midrule
    F+B & TF-IDF + Add    &            3.61 &            10.13 &            13.99 &            28.40 \\
    F+B & TF-IDF + Concat &            3.54 &             9.19 &            13.18 &            27.77 \\
    F+B & CLIP-B + Add          &            3.65 &             9.94 &            17.15 &            32.57 \\
    F+B & CLIP-B + Concat       &            3.58 &             9.70 &            16.88 &            32.49 \\
    F+B & CLIP-L + Add          &            \textbf{4.08} &            \textbf{10.95} &            \underline{17.20} &            \underline{33.03} \\
    F+B & CLIP-L + Concat       &            \underline{3.94} &            \underline{10.42} &            \textbf{17.49} &            \textbf{33.23} \\
    \midrule
    L+R & TF-IDF + Add    &            4.45 &            11.73 &            17.93 &            35.39 \\
    L+R & TF-IDF + Concat &            5.42 &            13.80 &            18.26 &            35.19 \\
    L+R & CLIP-B + Add          &            6.25 &            15.00 &            23.76 &            42.51 \\
    L+R & CLIP-B + Concat       &            \underline{8.02} &            \underline{18.61} &            \underline{25.14} &            42.82 \\
    L+R & CLIP-L + Add          &            6.41 &            15.52 &            24.76 &            \underline{43.74} \\
    L+R & CLIP-L + Concat       &            \textbf{8.36} &            \textbf{19.33} &            \textbf{26.66} &            \textbf{44.92} \\
    \midrule
    A & TF-IDF + Add           &            7.21 &            18.24 &            26.78 &            47.57 \\
    A & TF-IDF + Concat        &            7.75 &            18.63 &            26.37 &            46.08 \\
    A & CLIP-B + Add                 &            8.68 &            20.87 &            34.47 &            \underline{54.63} \\
    A & CLIP-B + Concat              &           \underline{10.50} &            \underline{23.82} &            33.79 &            52.92 \\
    A & CLIP-L + Add                 &            9.44 &            22.70 &            \textbf{35.46} &            \textbf{55.99} \\
    A & CLIP-L + Concat              &           \textbf{11.48} &            \textbf{25.27} &            \underline{35.23} &            54.49 \\
    \bottomrule
  \end{tabular}
\end{table}

\begin{table}[t]
  \caption{Ablation study of multiple sensors combinations for the proposed MSSPlace method. In the Modalities column, we denote modalities combinations, in brackets we specify the cameras used as the source for images, semantic masks and text descriptions. We use the following abbreviations: L -- LiDAR point clouds, I -- images, S -- semantic masks, T -- text descriptions, F -- front camera, A -- all cameras. 
  }
  \label{tab:ablation_modalities}
  \centering
  \scriptsize
  \begin{tabular}{lccccc}
    \toprule
    \multirow{2}{*}{Modalities} & Descriptor & \multicolumn{2}{c}{Oxford RobotCar} & \multicolumn{2}{c}{NCLT} \\
               & dimension & AR@1 & AR@1\% & AR@1 & AR@1\% \\
    \midrule
    I(F)             & 256 & 84.56 & 93.62 & 65.88 & 76.88 \\ 
    I(A)             & 256 & 89.84 & 97.23 & 88.50 & 94.75 \\
    I(A)+S(A)        & 512 & 89.85 & 97.23 & 88.50 & 94.74 \\
    I(A)+T(A)        & 512 & 89.90 & 97.24 & 84.79 & 93.09 \\
    I(A)+S(A)+T(A)   & 768 & 89.88 & 97.23 & 87.59 &94.37 \\
    L                & 256 & 94.68 & 98.04 & 91.87 & 95.92 \\
    L+I(F)           & 512 & 97.58 & 99.09 & 91.46 & 96.03 \\
    L+I(A)           & 512 & \underline{98.21} & \textbf{99.53} & \textbf{94.67} & \textbf{97.72} \\
    L+I(A)+S(A)      & 768 & 97.85 & \underline{99.47} & \underline{93.57} & \underline{97.16} \\
    L+I(A)+T(A)      & 768 & \textbf{98.22} & \textbf{99.53} & 92.36 & 96.51 \\
    L+I(A)+S(A)+T(A) & 1024 & 97.85 & \underline{99.47} & 92.51 & 96.51 \\
    \bottomrule
  \end{tabular}
\end{table}

\begin{table}[t]
  \caption{The comparative results of existing methods with the proposed MSSPlace approach on the Oxford RobotCar dataset. In the Sensors column, we denote sensor combinations and cameras used. We use the following abbreviations: L -- LiDAR point clouds, I -- images, S -- semantic masks, T -- text descriptions, F -- front camera, A -- all cameras. 
  }
  \label{tab:oxford_results}
  \centering
  \scriptsize
  \begin{tabular}{lcccc}
    \toprule
    \multirow{2}{*}{Method} & Descriptor & \multirow{2}{*}{Sensors} & \multirow{2}{*}{AR@1} & \multirow{2}{*}{AR@1\%} \\
     & dimension & & & \\
    \midrule
    NetVLAD \cite{arandjelovic_netvlad_2016} & 32768 & I(F)& 52.54&64.62\\
    CosPlace \cite{berton_rethinking_2022} & 256 & I(F)& 83.46&88.86\\
    MixVPR \cite{ali-bey_mixvpr_2023} & 4096 & I(F)& 88.68&92.60\\
    EigenPlaces \cite{berton2023EigenPlacesTrainingViewpoint} & 256& I(F)& 83.41&88.79\\
    AnyLoc \cite{keetha2024AnyLocUniversalVisual} & 49152 & I(F)& 82.94&91.25\\
    \midrule
    SVT-Net \cite{fan_svt-net_2022}              & 256 & L       & 93.7 &	97.8  \\
    ASVT-Net \cite{fan_svt-net_2022}             & 256 & L       & 93.9  & 98.0  \\
    MinkLoc3Dv2 \cite{komorowski_improving_2022} & 256 & L       & 96.3  & 98.9  \\
    \midrule
    CORAL \cite{pan_coral_2021}                  & 256& L+I(F) & 88.9  & 96.1  \\
    PIC-Net \cite{lu_pic-net_2020}               & 512& L+I(F) & -     & 98.2  \\
    AdaFusion \cite{lai_adafusion_2022}          & 256& L+I(F) & 98.18 & \underline{99.21} \\
    MinkLoc++ \cite{komorowski_minkloc_2021}     & 256 & L+I(F) & 97.15 & 99.06 \\
    \midrule
    MSSPlace-LI (ours)  & 512 & L+I(A)      & \underline{98.21} & \textbf{99.53} \\
    MSSPlace-LIS (ours) & 768 & L+I(A)+S(A) & 97.85 & 99.47 \\
    MSSPlace-LIT (ours) & 768 & L+I(A)+T(A) & \textbf{98.22} & \textbf{99.53} \\
    \bottomrule
  \end{tabular}
\end{table}

\begin{figure*}[t]
  \centering
  \includegraphics[width=0.99\linewidth]{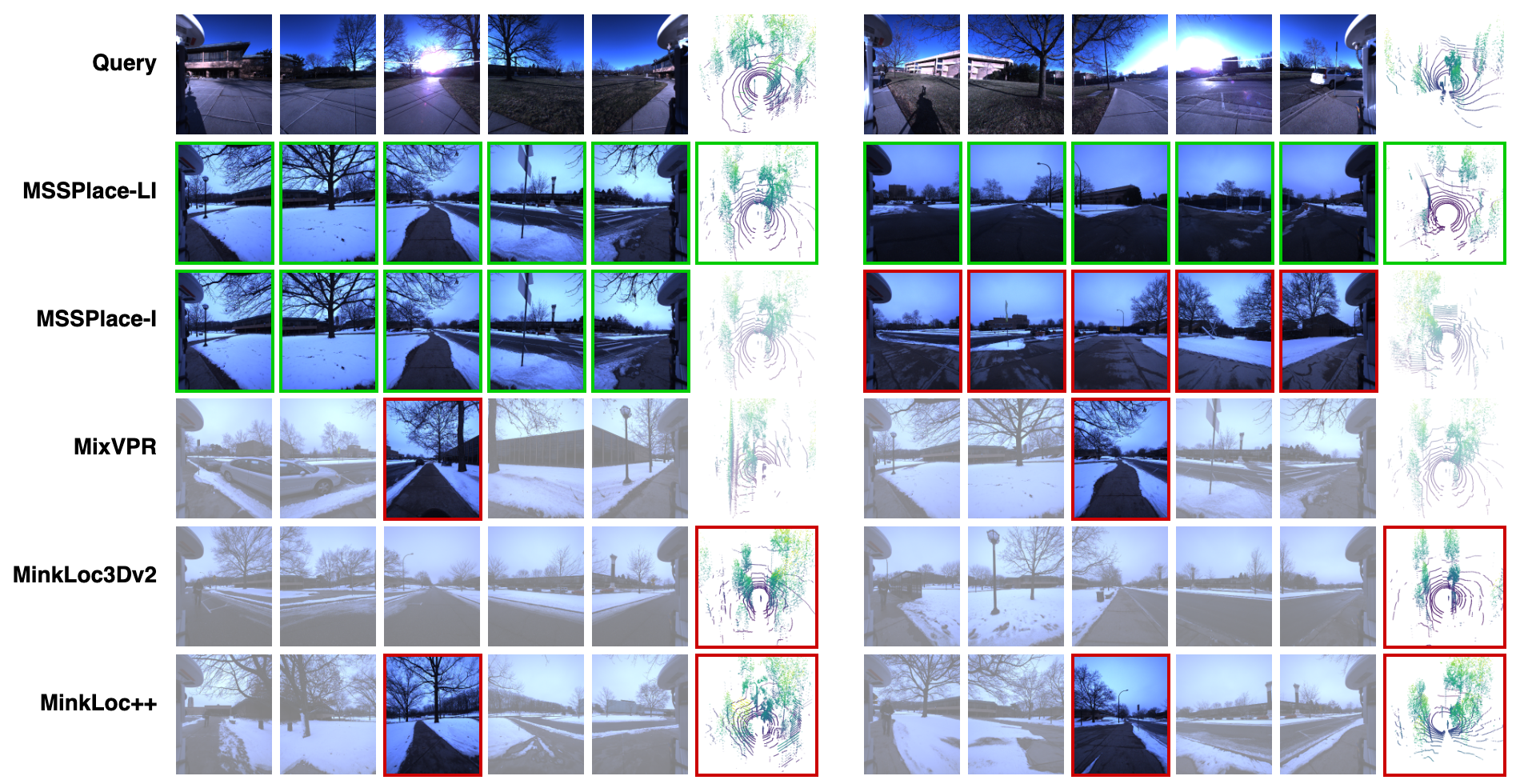}
  \caption{The visualization shows two samples and retrieval results obtained with different methods. 
  Correct matches are highlighted in green rectangles, erroneous results are highlighted in red.
  It can be observed that using additional cameras together with a LiDAR sensor may improve the performance of place recognition.}
  \label{fig:errors_vis}
\end{figure*}

\begin{table}[t]
  \caption{The comparative results of existing methods with the proposed MSSPlace approach on the NCLT dataset. In the Sensors column, we denote sensor combinations and cameras used. We use the following abbreviations: L -- LiDAR, I -- images, S -- semantic masks, T -- text descriptions; F -- front camera, A -- all cameras. 
  }
  \label{tab:nclt_results}
  \centering
  \scriptsize
  \begin{tabular}{lcccc}
    \toprule
    \multirow{2}{*}{Method} & Descriptor & \multirow{2}{*}{Sensors} & \multirow{2}{*}{AR@1} & \multirow{2}{*}{AR@1\%} \\
     & dimension & & & \\
    \midrule
    NetVLAD \cite{arandjelovic_netvlad_2016} & 32768& I(F)& 54.87& 67.47\\
    CosPlace \cite{berton_rethinking_2022} & 256& I(F)& 74.17& 79.75\\
    MixVPR \cite{ali-bey_mixvpr_2023} & 4096& I(F)& 75.25& 79.99\\
    EigenPlaces \cite{berton2023EigenPlacesTrainingViewpoint} & 256& I(F)& 74.18& 80.02\\
    AnyLoc \cite{keetha2024AnyLocUniversalVisual} & 49152& I(F)& 73.85& 83.42\\
    \midrule
    ASVT-Net \cite{fan_svt-net_2022}  & 256& L& 86.45&93.09\\
    SVT-Net \cite{fan_svt-net_2022} & 256& L& 79.24&89.35\\
    MinkLoc3Dv2 \cite{komorowski_improving_2022} & 256& L& 90.43&95.17\\
    \midrule
    AdaFusion \cite{lai_adafusion_2022}          & 256& L+I(F) & \textbf{95.65} & - \\
    MinkLoc++ \cite{komorowski_minkloc_2021} & 256& L+I(F) & 91.51&96.08\\
    \midrule
    MSSPlace-LI (ours)       & 512& L+I(A)     & \underline{94.67} & \textbf{97.72} \\
    MSSPlace-LIS (ours) & 768 & L+I(A)+S(A) & 93.57 & 97.16 \\
    MSSPlace-LIT (ours) & 768 & L+I(A)+T(A) & 92.36 & 96.51 \\
    \bottomrule
  \end{tabular}
\end{table}

\subsection{Experiments with multiple modalities}
\label{subsec:multimodal_experiments}

For ablation study of multimodal scenario, we combine the models trained for separate modalities to produce global place descriptors (see \Cref{tab:ablation_modalities}). We use the models with \textit{Add} fusion method for camera images and semantic masks, and \textit{CLIP-L+Add} models for text descriptions. For the fusion of multiple modalities, we opt for concatenation. 

The best result for the proposed MSSPlace approach is achieved in the configuration combining LiDAR and images from all cameras. This configuration significantly outperforms the setup with LiDAR and the front camera image alone, underscoring the utility of multiple cameras in enhancing performance.

The incorporation of semantic masks and text descriptions alongside camera images and LiDAR point clouds does not yield an improvement in the overall quality. These modalities, being derivatives of camera images, do not introduce novel information when combined with images, despite being useful by themselves.

The comparative results of existing methods with the proposed MSSPlace approach on the Oxford RobotCar and NCLT datasets are detailed in \Cref{tab:oxford_results} and \Cref{tab:nclt_results}. 


From \Cref{tab:oxford_results} it is clear that the proposed variants of the MSSPlace approach outperform existing both unimodal and multimodal methods on the Oxford RobotCar dataset, reaching 98.22\% average recall (top-1). On the NCLT dataset (see \Cref{tab:nclt_results}), our proposed MSSPlace approach achieves state-of-the-art results with 97.72\% average recall (top-1\%). It should be noted that for the AdaFusion approach, the results are from its original article due to the lack of source code to reproduce them.

The illustration of qualitative comparative results is demonstrated in \Cref{fig:errors_vis}.
It shows the superiority of the proposed MSSPlace method, which allows the use of multiple cameras and LiDAR to construct a high-quality vehicle place descriptor.

\section{Conclusion}
\label{sec:conclusion}

Our research has contributed to the field of place recognition by introducing a modular neural network method named MSSPlace. We systematically studied the performance of each modality independently, providing valuable insights into their individual strengths and limitations. Through comprehensive experiments, we explored various combinations of different modalities. Our experiments showed that the best performance may be achieved in the configuration with LiDAR and multiple cameras.

The utilization of semantic masks and text descriptors separately yielded promising results, but their combined use with image data either lacked a noticeable impact or degraded the quality of place recognition. This observation suggested that the image data inherently encompassed all necessary information, and the explicit representation of semantics did not contribute additional information.

An important feature of our method is its modular structure, which allows to substitute the neural network architecture for each modality, providing a flexible configuration. We believe that adopting more contemporary and sophisticated architectures for image and LiDAR-based place recognition can substantially improve the overall quality of our results.

One potential direction for future research is to investigate optimal descriptor sizes for each modality in order to capture their unique characteristics more effectively. By tailoring descriptor sizes to individual modalities, we can potentially enhance the discriminative power and information representation capability of multimodal place recognition systems.

\textit{Limitations:}
While our proposed approach has demonstrated promising results in the experiments conducted on the available datasets, it is important to acknowledge the limitations of our study. The scope of our claims is primarily based on the evaluation performed on a limited number of datasets and a specific number of training runs. It is essential to conduct further investigations with a larger and more diverse datasets to validate the generalization of our approach. 

\section*{Acknowledgment}

This work was partially supported by the Analytical Center for the Government of the Russian Federation in accordance with the subsidy agreement (agreement identifier 000000D730321P5Q0002; grant No. 70-2021-00138).


\ifCLASSOPTIONcaptionsoff
  \newpage
\fi



\bibliographystyle{IEEEtran}
\bibliography{main}
%



%

\begin{IEEEbiography}[{\includegraphics[width=1in,height=1.25in,clip,keepaspectratio]{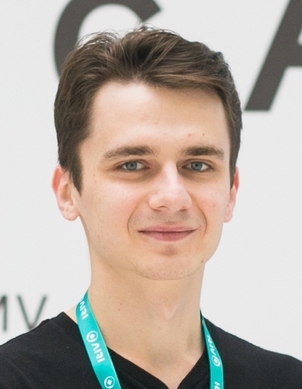}}]{Alexander Melekhin} received a Bachelor's degree in computer science in 2019 and a Master's degree in computer science in 2021 from the Moscow Technical University of Communications and Informatics.

Since 2021, he has been a research engineer at the Intelligent Transport Laboratory, Moscow Institute of Physics and Technology. Since 2022, he has been a Ph.D. student at the Moscow Institute of Physics and Technology. His research interests include computer vision and deep learning.
\end{IEEEbiography}

\begin{IEEEbiography}[{\includegraphics[width=1in,height=1.25in,clip,keepaspectratio]{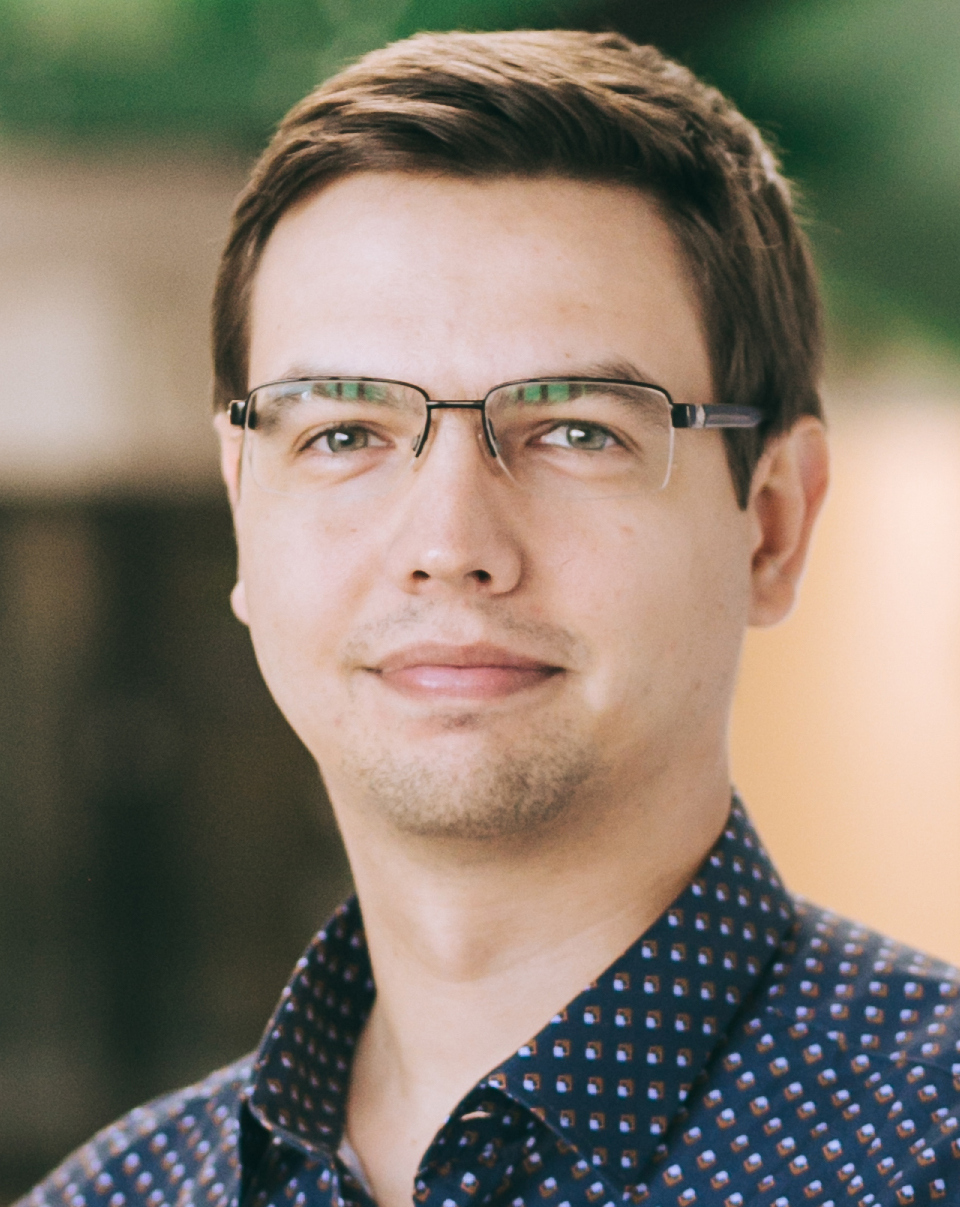}}]{Dmitry A. Yudin} received the engineering diploma in automation of technological processes and production in 2010 and the Ph.D. degree in computer science from the Belgorod State Technological University (BSTU) named after V.G. Shukhov, Belgorod, Russia in 2014.

From 2009 to 2019, he was a Researcher and Assistant Professor with Technical Cybernetics Department at BSTU n.a. V.G. Shukhov. Since 2019, he has been the head of the Intelligent Transport Laboratory at the Moscow Institute of Physics and Technology, Moscow, Russia. Since 2021, he has been a Senior Researcher at AIRI (Artificial Intelligence Research Institute), Moscow, Russia. 
He is the author more than 100 articles. His research interests include computer vision, deep learning, and robotics.  
\end{IEEEbiography}

\begin{IEEEbiography}[{\includegraphics[width=1in,height=1.25in,clip,keepaspectratio]{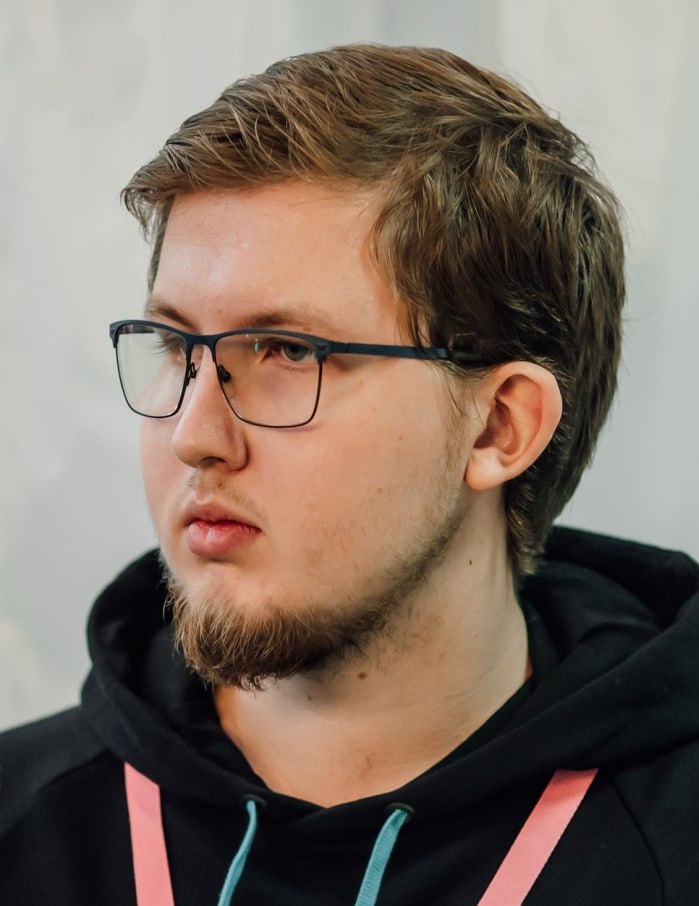}}]{Ilia Petryashin} was born in Cheboksary, Russia, in 2000. In 2018, he received a bachelor's degree in relay protection and automation from I.N. Ul'yanov Chuvash State University, and is currently a master's student in Methods and Technologies of Artificial Intelligence at Moscow Institute of Physics and Technology.

From 2022 to 2024, he is an engineer at the Intelligent Transport Laboratory, Moscow Institute of Physics and Technology, Moscow, Russia. Since 2024, he has been the data science engineer at Avito. Author of more than 30 articles and patents. His research interests are computer vision and deep learning.
\end{IEEEbiography}

\begin{IEEEbiography}[{\includegraphics[width=1in,height=1.25in,clip,keepaspectratio]{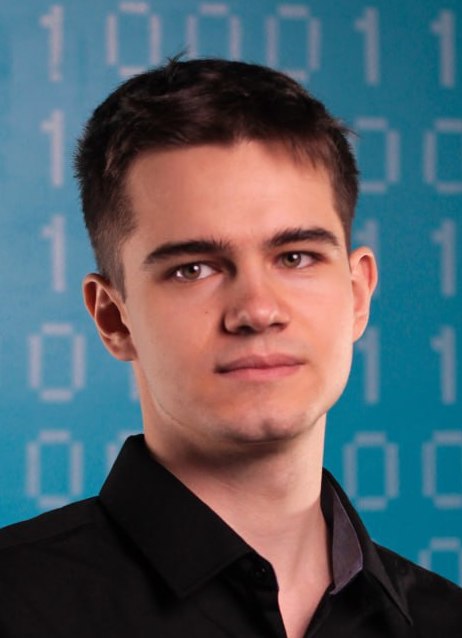}}]{Vitaly Bezuglyj} was born in Kiev, Ukraine, in 2000. In 2022 he received a Bachelor's degree in Applied Mathematics and Computer Science from the National Research Nuclear University MEPhI, Moscow, Russia.

From 2021 to 2023, he was working as a research engineer at the Intelligent Transport Laboratory, Moscow Institute of Physics and Technology, Moscow, Russia.  His research interests are three-dimensional semantic mapping, computer vision and robotics.
\end{IEEEbiography}




\end{document}